\documentclass[lettersize,journal]{IEEEtran}
\usepackage{amsmath,amsfonts}
\usepackage{algorithmic}
\usepackage{algorithm}
\usepackage{array}
\usepackage[caption=false,font=normalsize,labelfont=sf,textfont=sf]{subfig}
\usepackage{textcomp}
\usepackage{booktabs}
\usepackage{stfloats}
\usepackage{url}
\usepackage{verbatim}
\usepackage{multirow}
\usepackage{graphicx}
\usepackage{cite}
\usepackage{color}
\begin{document}

%
\title{Learning Using Privileged Information for Zero-Shot Action Recognition}
%
%
%

\author{Zhiyi Gao, Yonghong Hou,~\IEEEmembership{Member,~IEEE,} Wanqing Li,~\IEEEmembership{Senior~Member,~IEEE,} Zihui Guo and Bin Yu 
\thanks{Zhiyi Gao, Yonghong Hou and Zihui Guo are with the School of Electrical and Information Engineering, Tianjin University, Tianjin 300072, China.  (e-mail:
zhiyigao@tju.edu.cn; houroy@tju.edu.cn; gzihui@tju.edu.cn)}
\thanks{Wanqing Li is with the Advanced Multimedia Research Laboratory, University of Wollongong, Wollongong, NSW 2522, Australia (e-mail: wanqing@uow.edu.au)}
\thanks{Bin Yu is with Tianjin International Engineering Institute, Tianjin University, Tianjin 300072, 
China.(e-mail: yubin\_1449508506@tju.edu.cn)}}

\maketitle

\begin{abstract}
\label{sec:0.abstract}
Zero-Shot Action Recognition (ZSAR) aims to recognize video actions that have never been seen during training. Most existing methods assume a shared semantic space between seen and unseen actions and intend to directly learn a mapping from a visual space to the semantic space. This approach has been challenged by the semantic gap between the visual space and semantic space. This paper presents a novel method that uses object semantics as privileged information to narrow the semantic gap and, hence, effectively, assist the learning. In particular, a simple hallucination network is proposed to implicitly extract object semantics during testing without explicitly extracting objects and a cross-attention module is developed to augment visual feature with the object semantics. Experiments on the Olympic Sports, HMDB51 and UCF101 datasets have shown that the proposed method outperforms the state-of-the-art methods by a large margin.
\end{abstract}


%
\IEEEpeerreviewmaketitle

\section{Introduction}
\IEEEPARstart{R}{esearch} on action recognition from videos has made rapid progress in the past years with ceiling performance being reached on some datasets \cite{feichtenhofer2016convolutional,carreira2017quo, feichtenhofer2017spatiotemporal,wang2017spatiotemporal,zhu2020comprehensive,li2021trear,li2022self}. However, with the growing number of action categories, traditional supervised approach suffers from scalability problem. Moreover, annotating sufficient examples for the ever-growing new categories in real-world is cost-expensive and time-consuming. Therefore, extending a well-trained model to new/unseen classes, known as Zero-Shot Action Recognition (ZSAR), is gaining increasing interest recently. 

\begin{figure}[t]
  \centering
  \includegraphics[width=1.0\linewidth]{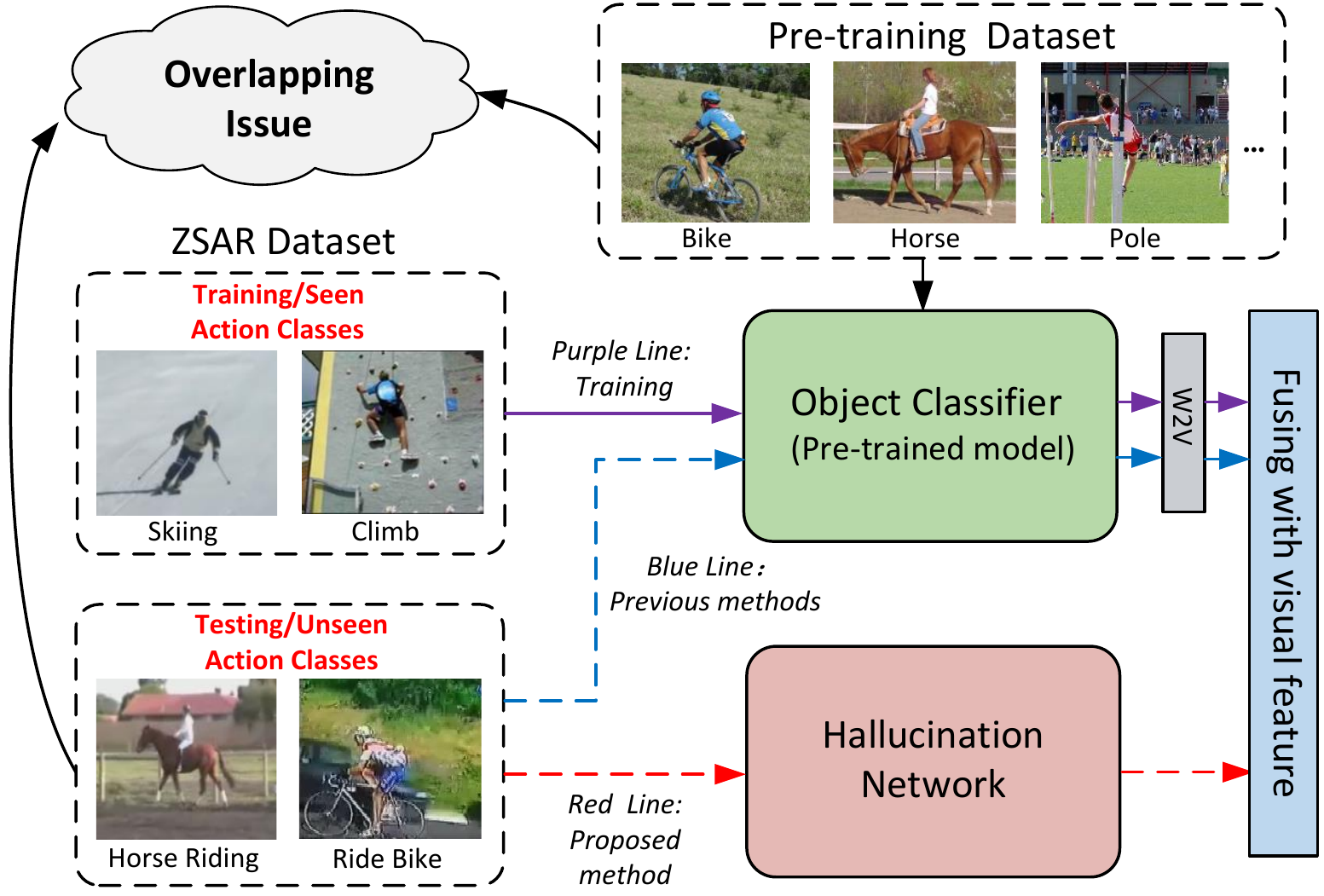}

   \caption{Proposed method vs existing methods. An object classifier pretrained on ImageNet is used to extract objects during training (purple line).  When testing, the unseen actions are high-related to ImageNet classes. The previous methods (blue line) still require an object classifier. Instead, the proposed method (red line) uses a hallucination network to extract object related information.}
   \label{ZSAR}
\end{figure}

In general, ZSAR assumes that both seen and unseen classes of actions share a common semantic space, e.g. a high dimensional vector space, and is achieved by learning a mapping from visual feature to the semantic space. Once the mapping is learned, an unseen sample is classified by the nearest neighbor search in the semantic space. There are a number of ways to define the semantic space. Typical early works \cite{liu2011recognizing,fu2014transductive,wang2017zero} define the semantic space using hand-crafted attributes. Recently, the semantic space is often defined as the word embedding space of action names, i.e. word2vec, using a pretrained language model \cite{mikolov2013efficient}. As for the mapping, direct learning from seen actions has been studied in \cite{brattoli2020rethinking}. However, visual observation and action names or hand-crafted attributes are two different modalities. The semantic gap between the two modalities has been challenging the effectiveness and robustness of learning such a mapping from visual feature to semantic representation. 

To mitigate the semantic gap, several approaches have been reported recently. One approach is to define a semantic space using example images \cite{wang2017alternative} e.g. considering the visual feature of the images as the semantic space. The second approach is to amend the visual feature with semantic information. For instance, in \cite{jain2015objects2action}, a pretrained object classifier is employed to recognize objects from action videos, and embedding of object names is treated as the visual representation without considering spatio-temporal information of the actions. In \cite{su2021vdarn}, objects are also extracted and the embedding of their names is concatenated to the feature representing poses to form the visual feature. Since such visual feature carries some amount of semantic information, the semantic gap is expected to be narrowed. 

It has to be pointed out that both\cite{jain2015objects2action}  and  \cite{su2021vdarn} require an object classifier to extract objects and word-embed their names during testing. The classifier is pretrained on a large dataset such as the ImageNet \cite{deng2009imagenet}, this practice has raised a question of validity of their methods being truly ZSAR because the large-scale dataset that is used to train the object classifier likely contains images high-related to unseen action classes (see Fig.~\ref{ZSAR}). For instance, images representing actions ``Ride Bike" and ``Horse Riding" are found in ImageNet \cite{deng2009imagenet} and these actions are likely to be considered as unseen actions in the random training-test split of ZSAR evaluation.

Nevertheless, using information of objects relevant to actions is a promising strategy. It not only narrows the semantic gap via word-embedding of the extracted object names, but also provides information in addition to the spatio-temporal visual information of the actions. As a true ZSAR, it is desirable that the object classifier pretrained on a dataset should not be employed during testing for the sake of computation and avoiding potential licenses fee in commercial applications.  This paper presents such a method based on the paradigm of learning using privileged information (LUPI) \cite{vapnik2009new}. Specifically, objects are annotated or extracted offline from seen actions and their names are word-embedded into a vector in the visual space as privileged information (PI) in training. Unlike the methods in \cite{jain2015objects2action} and \cite{su2021vdarn}, our method \textit{does not need the object classifier during testing phase}, instead it uses a hallucination network to mimic the extraction of related semantic information. The output of the hallucination network is fused with the visual feature by a cross-attention module to narrow the semantic gap and assist the mapping from visual feature to the semantic space.  Experiments on the widely used Olympic Sports \cite{niebles2010modeling}, HMDB51 \cite{kuehne2011hmdb} and UCF101 \cite{soomro2012ucf101} datasets show that the proposed method achieves the state-of-the-art results and outperforms the existing methods by a large margin.

The main contributions of this paper are as follows: 

\begin{itemize}
\item{A novel ZSAR method that implicitly leverages object information to narrow the semantic gaps and assists learning of a mapping from visual feature to a semantic space.}
\item{A hallucination network with the privileged information of objects related to actions to ``imitate" the object semantics during testing.}
\item{A cross-attention module to fuse visual feature with object semantics or the feature from the hallucination network. }
\item{Extensive evaluation on the widely used Olympic Sports, HMDB51 and UCF101 benchmarks achieves the state-of-the-art results.}
\end{itemize}

\begin{figure*}
\centering
\includegraphics[width=1.0\linewidth]{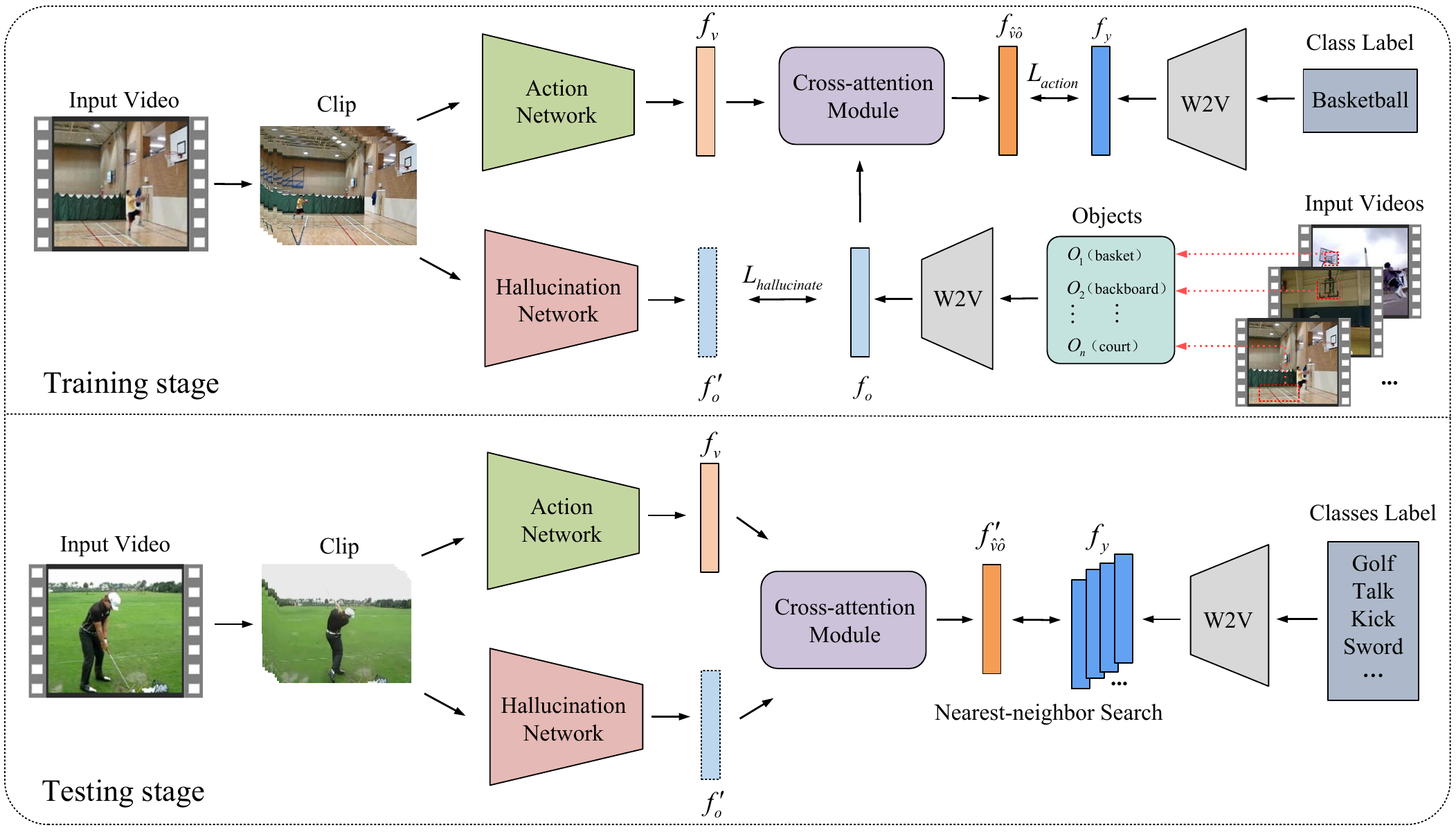}
\caption{The illustration of the proposed framework. It takes visual feature and the corresponding object semantics as input during training. A hallucination network is simultaneously trained to implicitly extract object related semantics and replace the object classifier at test time.  A cross-attention module is designed to fuse visual feature with object semantics or the feature from hallucination network. At test time, the nearest neighbor search is used for prediction in the semantic space. Gray blocks represent modules that are fixed during training. Colors (green, red, purple) blocks indicate modules trained.}
\label{fig:framework}
\end{figure*}
\section{Related work}
\label{sec:Related}

\subsection{Zero-Shot Action Recognition}
In general, ZSAR is achieved by defining a common semantic space for seen and unseen actions and learning a mapping between the visual space and semantic space. In relation to the semantic space, early works define it based on hand-crafted attributes \cite{liu2011recognizing,fu2014transductive,wang2017zero,gong2019zero}. However, the definition and annotation of hand-crafted attributes are subjective and labour-intensive. Recently, word-embedding of action names \cite{xu2015semantic,bishay2019tarn,brattoli2020rethinking} is favorably adopted.  In addition, textual descriptions of human actions or visual features extracted from still images are studied as a semantic space \cite{wang2017alternative}. In \cite{zhou2017semantic,zhou2019jointly}, semantically meaningful poses extracted from associated textual instructions of actions are also defined as a semantic space. Chen and Huang \cite{chen2021elaborative} use the textural description, called Elaborative Description (ED), of actions and embed them to define a semantic space. As for the visual space, it is generally defined through hand-crafted features \cite{lampert2009learning} (e.g. the Dense Trajectory Features (DTF) \cite{wang2013action} and the Improved Dense Trajectories (IDT) \cite{wang2016robust}) or deep features \cite{tran2018closer} extracted with the pre-trained 3D CNN-based Networks (e.g. Convolutional 3D Network (C3D) \cite{tran2015learning}, Inflated 3D Network (I3D) \cite{carreira2017quo}).
After constructing semantic space and visual space, most existing works directly learn a mapping from the visual space to semantic space \cite{xu2017transductive,brattoli2020rethinking}. However, this approach has been challenged by the semantic gap between the visual space and semantic space. Recently, some works attempt to amend the visual feature with object semantics in order to narrow the semantic gap between the visual space and semantic space. For example, Jain et al. \cite{jain2015objects2action} extract objects from action videos by a pretrained object classifier and embed object names to form a visual representation. In \cite{su2021vdarn}, the object information and pose information are simultaneously extracted to form visual feature. However, both of them utilize the pretrained object classifier at test time, which is questionable as illustrated in the introduction. In contrast, the proposed method in this paper does not use any object classifier during testing.

\subsection{Learning Using Privileged Information}
Vapnik and Vashist \cite{vapnik2009new} introduce the paradigm of learning using privileged information (LUPI). It assumes that there are additional data available during training, referred to as privileged information (PI), but not available in testing.
A number of approaches have been proposed for action recognition by using privileged information.
Niu et al. \cite{niu2015visual} utilize textual features extracted from the contextual descriptions of web images and videos as privileged information to train a robust action classifier. 
Motiian et al. \cite{motiian2016information} explore several types of privileged information such as motion information or 3D skeletons for improving model learning. 
Crasto et al. \cite{crasto2019mars} regard optical flow as privileged information, along with RGB for training, but only RGB is used in test to avoid flow computation. 
Similarly, Garcia et al. \cite{garcia2020learning} consider that depth data is not going to be always available when a model is deployed in real scenarios. Therefore, depth data is used as privileged information along with RGB data during training, while employing only RGB data at test time. 
Different from the above methods, the proposed method uses object semantics from seen actions as privileged information to assist learning visual feature as well as a mapping between the visual feature and semantic space. In addition, a hallucination network is proposed to implicitly extract semantics during testing instead of relying on a pretrained object classifier.
\section{Method}
\label{sec:Method}
\subsection{Overview}
Let the training set $D_{s}=\{(v_{s}, y_{s})| v_{s} \in \mathcal{V}_{s}, y_{s} \in \mathcal{Y}_{s}\}$ consist of videos $v_{s}$ in video space $\mathcal{V}_{s}$ with labels from seen action classes $\mathcal{Y}_{s}$. Similarly, $D_{u}=\{(v_{u}, y_{u})| v_{u}\in\mathcal{V}_{u}, y_{u}\in\mathcal{Y}_{u}\}$ is the test set, which consists of videos $v_{u}$ with labels from unseen action classes $\mathcal{Y}_{u}$. Here, seen and unseen action classes are disjoint, i.e. $\mathcal{Y}_{s}\, \cap\, \mathcal{Y}_{u}=\emptyset$. The task of ZSAR is to train a model only on $D_{s}$, and predict the class labels of the unseen videos $v_{u}$.  The proposed architecture is illustrated in Fig.~\ref{fig:framework}, where the embedding space $f_{y}$ of action names is used as the shared semantic space, visual feature $f_{v}$ extracted by the action network is used as visual space. 
In addition, the object semantic $f_{o}$ for each known action class is utilized as privileged information to assist learning of a mapping from visual feature to the semantic space. The object semantic $f_{o}$ is the word embedding of object $o$ which may be obtained off-line using an object classifier. 
Specifically, in the training stage, video features $f_{v}$ and object semantics $f_{o}$ are fed to a cross-attention module and fused into a joint representation $f_{\hat{v} \hat{o}}$ being matched with the semantic representations $f_{y}$ in the semantic space. Meanwhile, a hallucination network is trained to “imitate” the object semantics $f_{o}$. At test time, the trained hallucination network is used to implicitly extract object related semantic $f^{\prime}_{o}$, which is fused with the visual feature $f_{v}$ by the cross-attention module to obtain the joint representation $f^{\prime}_{\hat{v} \hat{o}}$ as well. Finally, nearest-neighbor search method is used in the semantic space to recognize unseen action class. 

In the rest of the section, the five key components of the proposed framework including video and semantic embedding, object embedding, hallucination network, cross-attention module, and training and testing of the framework will be described in detail.

\subsection{Video and Semantic Embedding}

\subsubsection{Video embedding.} Given a video clip $v \in \mathbb{R}^{T \times H \times W}$ of $T$  frames with a height and width of $H$ and $W$ respectively, the action network consists of a pretrained R(2+1)D \cite{brattoli2020rethinking} network followed by a fully connected layer for extracting a video spatio-temporal feature $f_{v} \in \mathbb{R}^{300}$  

\subsubsection{Semantic embedding.} Following the majority of works \cite{xu2015semantic}, word2vec is applied to embed action names as semantic representation. Specifically, the skip-gram language model \cite{mikolov2013efficient} trained on 100 billion words from Google News articles is used to map each word into a 300 dimensional semantic space. For the class name containing multiple words, the vectors of all words are averaged to obtain the semantic embedding. That is, a class name consisting of $n$ words  $y=\left\{y_{1}, \cdots, y_{n}\right\}$ can be embedded as $f_{y}=\frac{1}{n}\sum_{i=1}^{n} \mathrm{~W} 2 V\left(y_{i}\right) \in \mathbb{R}^{300}$.

\subsection{Object Embedding as PI}
Objects are extracted by a pretrained object classifier from all video clips with $t$ frames uniformly sampled, and the top $k$ objects with the highest probabilities are reserved.
In order to associate the extracted objects with action classes, we calculate the frequency of the objects extracted from all clips of the same action class and preserve the top $m$ objects  $o=\left\{o_{1}, \cdots, o_{m}\right\}$. Then the objects are word-embedded as the privileged information of individual action class.  Similarly, the object semantic of each action class can be presented by $f_{o}=\frac{1}{m}\sum_{i=1}^{m} \mathrm{~W} 2 V\left(o_{i}\right) \in \mathbb{R}^{300}$.

\subsection{Hallucination Network}
\label{Hallucination}
Since the ImageNet dataset from which the object classifier is pretrained contains highly related categories with unseen action classes, we argue that the object classifier should not be employed during testing phase for truly ZSAR. To replace the object classifier, we introduce a hallucination network, which is trained to ``imitate” the object semantics. 
Thus at test time, it can extract the object related semantics $f^{\prime}_{o}$ to fuse with visual features. The hallucination network consists of a fixed pretrained R(2+1)D \cite{brattoli2020rethinking} network and four fully connected layers.

\subsection{Cross-attention Module}
\begin{figure}[t]
  \centering
  \includegraphics[width=0.8\linewidth]{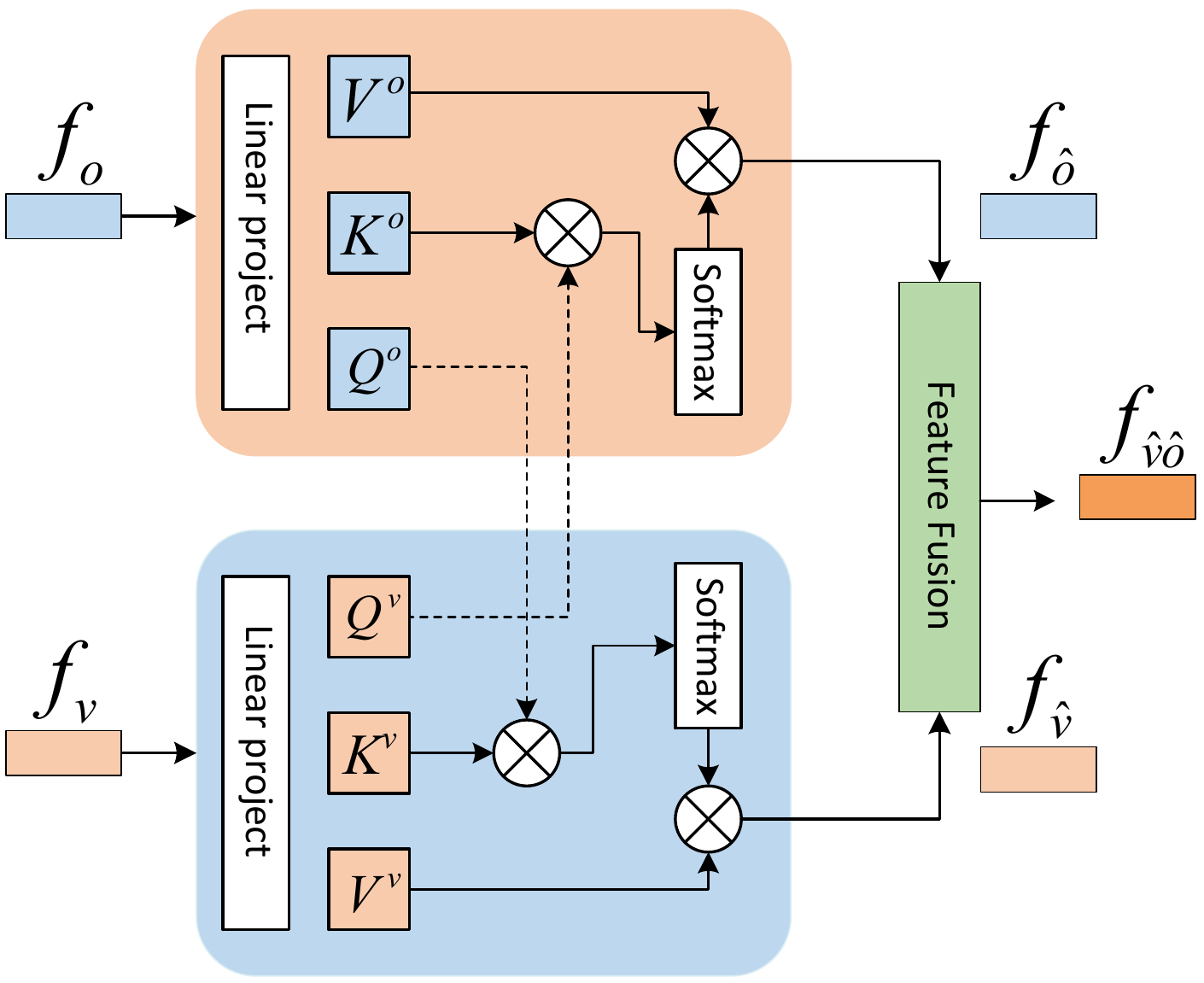}

   \caption{Illustration of the proposed cross-attention module. It consists of a mutual-attention layer and feature fusion operation. Visual features and object semantics or features from hallucination network are fed to the attention layer to exchange the information. Then the features are fused together to obtain the joint representations.}
   \label{fig:cross}
\end{figure}

We propose a cross-attention module to fuse visual feature $f_{v}$ with object semantics $f_{o}$ (training stage) or the feature $f^{\prime}_{o}$ (testing stage) from the hallucination network using mutual-attentional mechanism. As shown in Fig.~\ref{fig:cross}, the proposed module contains two parts, mutual-attention layer and feature fusion operation. Take the training phase as an example, mathematically, visual feature $f_{v}$  and object semantic $f_{o}$ are first mapped to a series vectors query ($Q^{v}$/$Q^{o}$) and key ($K^{v}$/$K^{o}$) of dimension $d_{k}$, and value (${V}^{v}$/${V}^{o}$) of dimension $d_{v}$ using unshared learnable linear projections. Then the mutual-attention is applied to retrieve the information from  vectors (key $K^{v}$ and value ${V}^{v}$) of visual feature related to query vector $Q^{o}$ of object semantics and vice the verse. The dot-product of the $Q^{o}$ vector and $K^{v}$ vector goes through a softmax function to obtain the attention weights, and have a weighted sum of $V^{v}$. The process can be expressed as 
\begin{equation}\label{three}
f_{\hat{v}}=\operatorname{Softmax}\left(\frac{Q^{o} (K^{v})^{T}}{\sqrt{d_{k}}}\right){V}^{v}
\end{equation}
Similarly, we can calculate the attention weights on each object semantic with respect to visual feature to obtain $f_{\hat{o}}$.

\begin{equation}\label{three}
f_{\hat{o}}=\operatorname{Softmax}\left(\frac{Q^{v} (K^{o})^{T}}{\sqrt{d_{k}}}\right){V}^{o}
\end{equation}
Such a mutual-attention scheme takes cross-modal interactions into consideration. It calculates the visual feature attention in semantic modality and object semantic attention in visual modality, and produces corresponding cross-modal features. After mutual-attention layer, features from both modalities are fused via simply feature addition operation.

\begin{equation}\label{three}
f_{\hat{v} \hat{o}}=f_{\hat{v}}+f_{\hat{o}}
\end{equation}

\subsection{Training and Testing}
As illustrated in Fig.~\ref{fig:framework}, in the training stage, the visual feature  $f_{v}$ and object semantic $f_{o}$ are fused to obtain a joint representation $f_{\hat{v} \hat{o}}$ in the semantic space to match their equivalent semantic representation $f_{y}$. The loss function $\mathcal{L}_{action}$ is defined as

\begin{equation}\label{three}
\mathcal{L}_{action}= \left\|f_{y}- f_{\hat{v} \hat{o}}\right\|_{2}^{2}
\end{equation}
In addition, a regression loss between the object semantics $f_{o}$ and feature $f^{\prime}_{o}$ from the hallucination network is defined

\begin{equation}\label{three}
\mathcal{L}_{hallucinate}= \left\|f_{o}- f^{\prime}_{o}\right\|_{2}^{2}
\end{equation}
The full model is end-to-end trained, and the overall loss of the proposed model is defined as

\begin{equation}\label{three}
\mathcal{L}= \mathcal{L}_{action}+ \mathcal{L}_{hallucinate}
\end{equation}

At test time, a test video goes through both the action network and the hallucination network to produce the visual feature $f_{v}$ and object related semantics $f^{\prime}_{o}$. After the cross-attention module, a joint representation $f^{\prime}_{\hat{v} \hat{o}}$  is obtained. Finally, nearest-neighbor search is used in the semantic space to predict the label $\mathcal{P}$ of unseen action class as
\begin{equation}\label{three}
\mathcal{P}=\underset{y \in D_{u}}{\operatorname{argmin}} \cos (f^{\prime}_{\hat{v} \hat{o}}, f_{y})
\end{equation}
where $\cos (\cdot)$ is the cosine distance.

\begin{table*}
  \centering
	\renewcommand\tabcolsep{12.5pt}
	\caption{ZSAR performances on the three existing benchmarks compared with state-of-the-art methods.  VE and SE represent Visual Embedding and Semantic Embeding. FV represents Fisher Vectors, obj represents objects.  A represents the human annotated attribute vectors, W represents the word2vec embedding, and ED represents elaborative description. The average Top-1 accuracy ($\%$) ± standard deviation is reported.}
    \begin{tabular}{ccccccc}
    \toprule 
    Methods & Reference & VE  & SE  & Olympic Sports & HMDB51& UCF101 \\
    \midrule
	 O2A\cite{jain2015objects2action}   & ICCV 2015 & Obj    & W   & N/A & 15.6 & 30.3 \\
    MTE\cite{xu2016multi}   & ECCV 2016 & FV    & W   & 44.3 ± 8.1 & 19.7 ± 1.6 & 15.8 ± 1.3 \\
    ASR\cite{wang2017alternative}  & ECML-PKDD 2017 & C3D   & W   & N/A   & 21.8 ± 0.9 & 24.4 ± 1.0 \\
    GMM\cite{mishra2018generative}   & WAC V2018 & C3D   & W   & N/A   & 19.3 ± 2.1 & 17.3 ± 1.1 \\
    UAR\cite{zhu2018towards}   & CVPR 2018 & FV    & W   & N/A   & 24.4 ± 1.6 & 17.5 ± 1.6 \\
    CEWGAN\cite{mandal2019out}  & CVPR 2019 & I3D   & W     & 50.5 ± 6.9 & 30.2 ± 2.7 & 26.9 ± 2.8 \\
    TS-GCN\cite{gao2019know} & AAAI 2019 & GCN   & W   & 56.5 ± 6.6 & 23.2 ± 3.0 & 34.2 ± 3.1 \\
	 BD-GAN\cite{mishra2020zero}   & NEUROCOMPUTING 2020& C3D & A  & 49.80 ± 10.72   & 23.86 ± 2.95 & 18.73 ± 3.73 \\
    E2E\cite{brattoli2020rethinking}   & CVPR 2020 & R(2+1)D & W  & N/A   & 32.7 & 48 \\
    VDARN\cite{su2021vdarn}    & AD HOC NETW 2021  & GCN   & W   & 57.6 ± 3.4 & 21.6 ± 2.8 & 26.4 ±2.8 \\
    ER\cite{chen2021elaborative}    & ICCV 2021 & TSM   & ED    & 60.2 ± 8.9 & 35.3 ± 4.6 & 51.8 ± 2.9 \\
	\midrule
    Ours  &       & R(2+1)D & W   &   \textbf{61.9 ± 7.8}    & \textbf{38.8 ± 4.6} & \textbf{52.6 ± 2.4} \\
    \bottomrule
    \end{tabular}%
	
  \label{sota}%
\end{table*}%

\begin{table*}
  \centering
  \renewcommand\tabcolsep{7.2pt}
\caption{Ablation study of the proposed object semantics and hallucination network on Olympic Sports, HMDB51 and UCF101 datasets.  ``Baseline'' refers to directly mapping visual features to semantic space; ``Object Semantics''refers to extracting object semantics through object classifier to fuse with visual features during training, but the classifier is not used in the test stage; ``Hallucination Network'' refers to implicitly extracting object semantics to fuse with visual features at test time. The average Top-1 and Top-5 accuracy ($\%$) ± standard deviation are reported.}
    \begin{tabular}{ccccccccc}
    \toprule 
    \multirow{2}[4]{*}{Baseline} & \multirow{2}[4]{*}{Object Semantics} & \multirow{2}[4]{*}{Hallucination Network} & \multicolumn{2}{c}{Olympic Sports} & \multicolumn{2}{c}{HMDB51} & \multicolumn{2}{c}{ UCF101 } \\
\cmidrule(lr){4-5} \cmidrule(lr){6-7}  \cmidrule(lr){8-9}         &       &       & Top-1  & Top-5 & Top-1  & Top-5 & Top-1  & Top-5 \\
    \midrule
    \checkmark     & $\times$     & $\times$     & 52.8 ± 8.6 & 88.8 ± 3.8 & 32.7 ± 3.0 & 56.3 ± 3.4 & 47.9 ± 2.2 & 72.4 ± 3.4 \\
    \checkmark     & \checkmark     & $\times$    &  57.0 ± 8.3 &  91.2 ± 6.9 & 36.7 ± 4.3 & 65.6 ± 5.0 & 49.8 ± 2.5 & 76.9 ± 2.7 \\
    \checkmark    & \checkmark    & \checkmark    &  \textbf{61.9 ± 7.8} & \textbf{93.3 ± 6.4} & \textbf{38.8 ± 4.6} & \textbf{66.9 ± 3.7} & \textbf{52.6 ± 2.4} & \textbf{77.1 ± 2.9} \\
    \bottomrule 
    \end{tabular}%
	
  \label{ablation1}%
\end{table*}%

\section{Experiments}
\subsection{Setup}

\subsubsection{Datasets and Splits.} 
The proposed method is extensively evaluated on the Olympic Sports \cite{niebles2010modeling}, HMDB51 \cite{kuehne2011hmdb} and UCF101 \cite{soomro2012ucf101} datasets. The Olympic Sports dataset consists of 783 videos, which are divided into 16 sports actions. The HMDB51 dataset contains 6766 realistic videos distributed in 51 actions. The UCF101 dataset has 101 action classes with a total of 13,320 videos, which are collected from YouTube. To compare the proposed method with the state-of-the-art methods, we follow the same 50/50 data splits as used in  \cite{xu2017transductive}, i.e., videos of 50$\%$ categories are used for training and the rest 50$\%$ categories are considered as the unseen for testing. Specifically, experiments are conducted on the cases of  8/8, 26/25 and 51/50 splits for Olympic Sports, HMDB51 and UCF101, respectively. For each case, 30 independent splits for each dataset \cite{mishra2018generative} are randomly generated and the average accuracy and standard deviation are reported for experimental evaluation. 

\subsubsection{ZSAR Settings.} There are two ZSAR settings: inductive setting and transductive setting. The former assumes that only the labeled videos from the seen categories are available during training while the latter can use the unlabeled data of the unseen categories for model training. Specifically, in this work, we focus on inductive ZSAR \cite{brattoli2020rethinking,mishra2020zero,chen2021elaborative,su2021vdarn} and do not discuss the transductive approach \cite{fu2014transductive,xu2017transductive}.

\subsection{Implementation Details}

In the experiments, we use the pretrained R(2+1)D network \cite{brattoli2020rethinking} followed by a fully connected layer to extract spatio-temporal features. The object classifier is the BiT image model \cite{kolesnikov2020big} pretrained on ImageNet21k \cite{deng2009imagenet}. When extracting objects, frame $t$ of each video clip is set to 8 and top $k$ of all action clips is set to 20, top $m$ of each action class is set to 5. Both the action class and extracted objects are embedded using skip-gram word2vec model \cite{mikolov2013efficient}, which is pretrained on the Google News dataset. The word2vec model generates a 300-dimensional word vector representation for each word.  
For the input videos, we select a clip of 16 frames from each video, uniformly sampled in time, and each frame is cropped to 112 pixels × 112 pixels. The experiments are conducted on two TitanX GPUs. The networks are trained for 10 epochs on HMDB51 and UCF101 datasets, 20 epochs on the Olympic Sports dataset with a batch size of 16. 
\begin{table}[!t]
  \centering
  \renewcommand\tabcolsep{3.0pt}
\caption{Ablation study of the proposed cross-attention module with different fusion manners on HMDB51 and UCF101 datasets. The average Top-1 and Top-5 accuracy ($\%$) ± standard deviation are reported.}
    \begin{tabular}{ccccc}
    \toprule 
    \multirow{2}[4]{*}{Methods} & \multicolumn{2}{c}{HMDB51} & \multicolumn{2}{c}{ UCF101 } \\

\cmidrule(lr){2-3}  \cmidrule(lr){4-5}        & Top-1  & Top-5 & Top-1  & Top-5 \\
    \midrule
	 Multiplication &  34.1 ± 4.6  &  56.6 ± 3.6     &  49.4 ± 3.4  & 73.2 ± 3.6 \\
    Concatenation &  35.3 ± 4.7  &  58.6 ± 5.4  &   51.8+-2.8  & 75.4+-3.4 \\
    Addition & 37.1 ± 4.3 & 64.5 ± 4.4  &   51.9 ± 2.5  & 76.9 ± 3.5 \\
    Cross-attention & \textbf{38.8 ± 4.6} & \textbf{66.9 ± 3.7}& \textbf{52.6 ± 2.4} & \textbf{77.1 ± 2.9} \\
    \bottomrule 
    \end{tabular}%
  \label{ablation2}%
\end{table}%
We use Adam optimizer \cite{kingma2014adam} to train all networks. 
The learning rate is initially set to 1e-4 and is decayed by a factor of 0.5 every 5 epochs. The average Top-1 and Top-5 accuracy ($\%$) ± standard deviation are reported.

\subsection{Comparison with the State-of-the-art}
\label{State-of-the-art}
The comparison results are shown in Table~\ref{sota}. Overall, the proposed method performs best against state-of-the-art methods on three widely used datasets \cite{niebles2010modeling,kuehne2011hmdb,soomro2012ucf101}.  When using the same word embedding methods, our approach outperforms recent TS-GCN \cite{gao2019know}, and VDARN \cite{su2021vdarn} methods on three datasets. 
Compared with other attributes-based \cite{mishra2020zero} and ED-based \cite{chen2021elaborative} methods which are not scalable and labour-intensive, the proposed method uses the word embedding of action names as semantic space and still obtains better performance than them. 
Compared with E2E \cite{brattoli2020rethinking} method which also uses the R(2+1)D network and directly maps from visual features to the semantic space, our method improves by 6.1 and 4.6 percentage points on the HMDB51 and UCF101 datasets respectively, indicating that amending the visual feature with semantic information can effectively narrow the semantic gap.
Note that the O2A \cite{jain2015objects2action} and the VDARN \cite{su2021vdarn} methods also use object information as part of the visual feature, but they use object classifier during testing as well. Our method uses a hallucination network to extract object related information during testing and still performs better than them. 
In addition, the proposed method not only outperforms the very recent method ER \cite{chen2021elaborative} in classification accuracy, but also has a smaller standard deviation, which indicates that our method has a relatively stable performance under different training and testing data partitions.

\begin{figure}[t]
  \centering
  \includegraphics[width=1.0\linewidth]{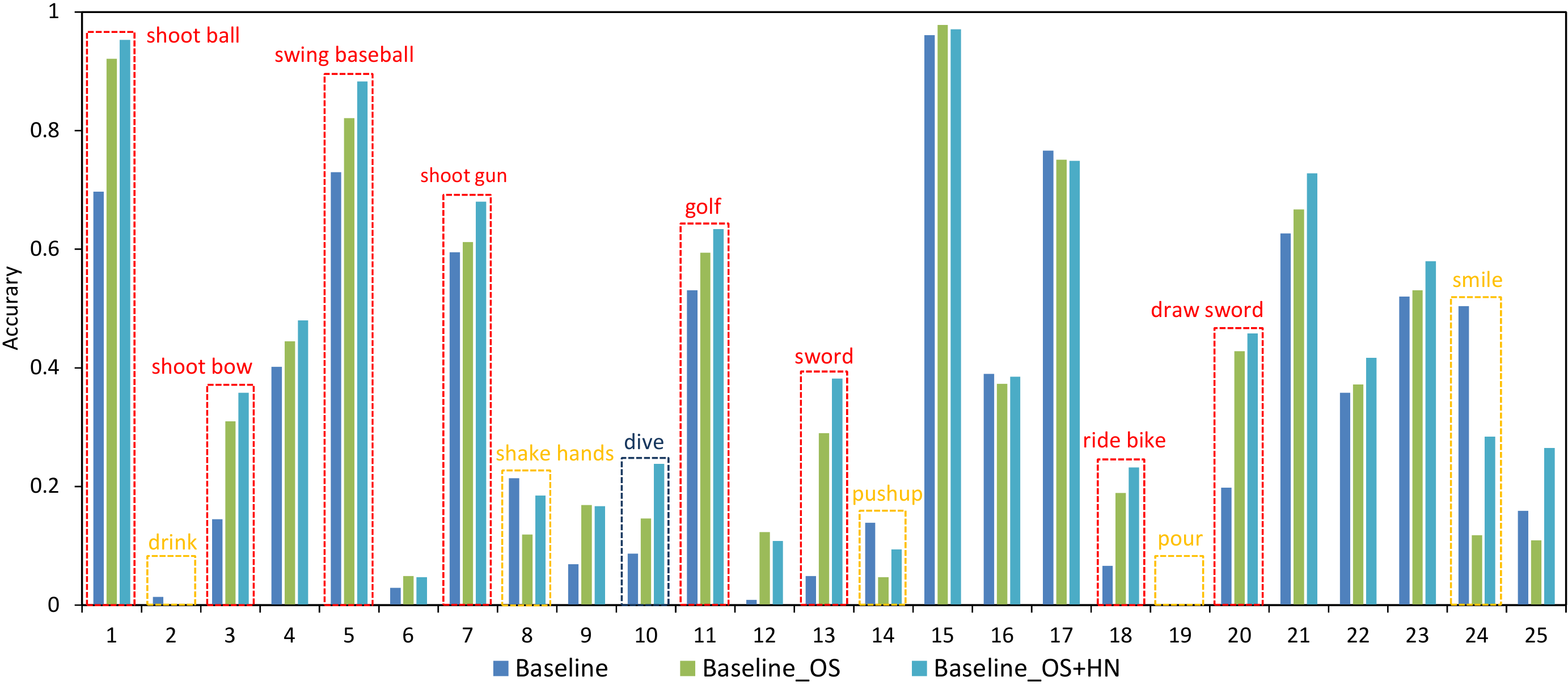}

   \caption{Accuracy comparison of each unseen action class on the HMDB51 dataset between the proposed method and baseline. Here, OS represents Object Semantics, HN represents Hallucination Network. The experimental settings of OS and HN are the same as Table~\ref{ablation1}. }
   \label{classes}
\end{figure}
\subsection{Ablation Studies}
\subsubsection{Impact of Object Semantics and Hallucination Network} 
\label{study1}To verify the effectiveness of the proposed object semantics and hallucination network, ablation studies are conducted on three benchmarks \cite{niebles2010modeling,kuehne2011hmdb,soomro2012ucf101}. As shown in the second row of Table~\ref{ablation1}, leveraging object semantics to augment visual feature only at the training stage can significantly improve classification accuracy on the three datasets. The hallucination network contributes further improvement by 4.9/2.1/2.8 percentage points for Top-1 accuracy and 2.1/1.3/0.2 percentage points for Top-5 accuracy on the three datasets respectively (see the last row of Table~\ref{ablation1}). Fig.~\ref{classes} further shows the accuracy comparison of each unseen action. It can be seen that the recognition accuracy of our method is significantly higher than the baseline in the red-marked categories, such as ``shooting ball'', ``shooting bow'', ``sword'', ``ride bike'' etc. These actions involve interaction with objects thus the hallucination network extracts the semantics of the related object to augment the visual feature and improve the recognition.  In addition, some actions such as ``dive'' (marked in blue) occur in a specific scene, and information on the scene is extracted by the hallucination network to augment the visual information. However, for some actions such as  ``smile'' and ``shake hands'' (marked in yellow), these actions do not involve any interaction with the surrounding environment. In this case, the information extracted by the hallucination network may interfere with the visual feature. The interference can be minimized by the choice of the objects in the training (See below section~\ref{Object Number} Impact of Object Number).


\begin{figure}[t]
  \centering
  \includegraphics[width=0.75\linewidth]{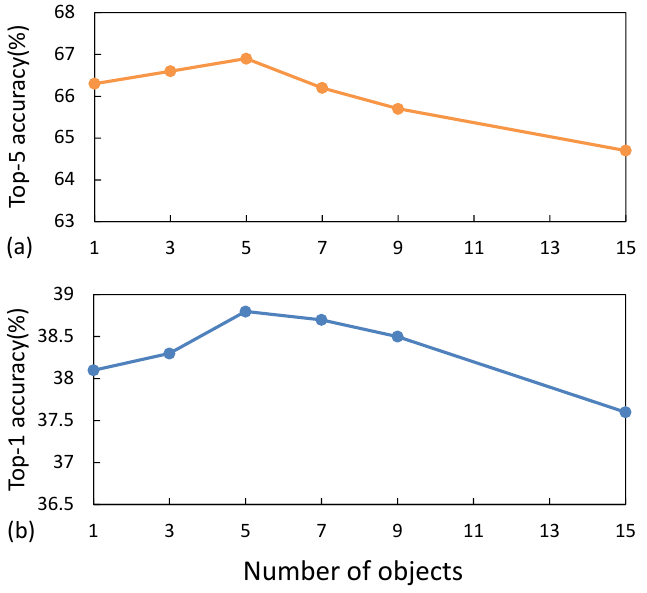}

   \caption{Evaluate the proposed method with different number of objects on HMDB51 dataset. The average Top-5 (a) and Top-1 (b) accuracy ($\%$) are illustrated.}
   \label{object5}
\end{figure}

\subsubsection{Impact of Fusion Strategy} In order to verify that the proposed cross-attention module is better than other fusion strategies, we perform another ablation experiment on the HMDB51 and UCF101 datasets, the results are shown in Table~\ref{ablation2}. Compared with multiplication, concatenation and addition fusing schemes, the proposed fusion strategy achieves the best performance. The reason is that other fusion strategies neglect the modality discrepancy. The proposed cross-attention block can effectively mitigate such an issue, in which the visual features and object semantics can exchange information through the mutual-attention layer to reduce the feature variations. Then the cross-modal features are fused to produce the conjoint feature representation. The results also show that the proposed cross-attention module has the smallest standard deviation on the UCF101 dataset.

\subsubsection{Impact of Object Number}
\label{Object Number}
Fig.~\ref{object5} shows the impact of object number used for each action class. It can be seen that the performance first increases as the number of objects increases and then decreases when the number of objects is over five for the HMDB51 dataset. This is probably due to that each action class is related to only a limited number of objects, and irrelevant objects will become noise interference and degrade performance. However, the performance degradation is insignificant (only about 0.3 percentage points) even if the object number is increased to nine.

\subsection{Qualitative Results}

\begin{figure}[t]
  \centering
  \includegraphics[width=1.0\linewidth]{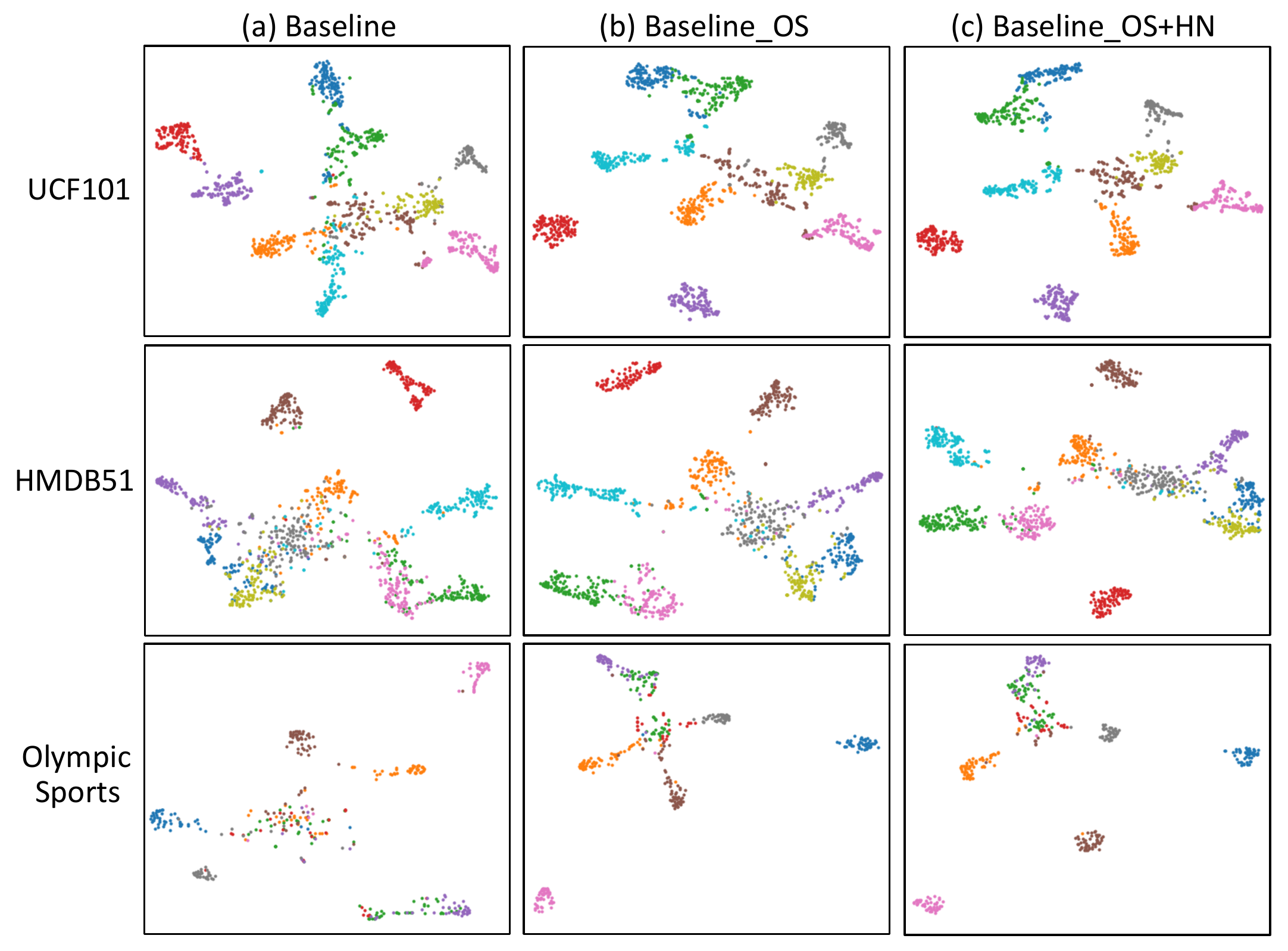}

   \caption{The t-SNE visualization of the visual feature distribution of unseen classes of the three datasets. Different classes are shown as dots in different colors. Here, OS represents Object Semantics, HN represents Hallucination Network. The experimental settings of OS and HN are the same as section~\ref{study1}.}
   \label{claster}
\end{figure}

We further analyze the effectiveness of the proposed method via the qualitative visualizations of visual feature distribution in the semantic space \cite{van2008visualizing}, as shown in Fig.~\ref{claster}. For the sake of visualizations, we randomly sample 8 unseen classes from the Olympic Sports dataset and 10 unseen classes from  HMDB51 and UCF101 datasets. As can be seen from Fig.~\ref{claster}(a) and Fig.~\ref{claster}(b), with the aid of the object classifier during training, the baseline network yield improved feature representations. In Fig.~\ref{claster}(c), we further use a hallucination network to mimic the object semantics to fuse with visual features at test time, which can obtain the tightest intra-class clusters and separable inter-class clusters in the semantic space. The visualization results prove that the proposed method can effectively use object semantics to amend visual representation, thus alleviating the semantic gap.




\section{Conclusions}
{In this paper, we propose a novel method for ZSAR. It is built upon the paradigm of learning using privileged information (LUPI). Specifically, the object semantics are used as privileged information to narrow the semantic gaps and assist learning of a mapping from visual feature to a semantic space. A hallucination network is trained to implicitly extract object related semantics at test time instead of using a pretrained object classifier. Moreover, a cross-attention module is designed to fuse visual feature with object semantics or the feature from hallucination network. Empirical experiments on the Olympic Sports, HMDB51 and UCF101 benchmarks have shown that the proposed method achieves the state-of-the-art results.}
\label{sec:reference}
\bibliographystyle{IEEEtran}
\bibliography{reference}

\end{document}